# Neural networks with image recognition by pairs

Geidarov P.Sh.

Institute of Control Systems NAS of Azerbaijan,

plbaku2010@gmail.com

Abstract

Neural networks based on metric recognition methods have a strictly determined architecture. Number of neurons, connections, as well as weights and thresholds values are calculated analytically, based on the initial conditions of tasks: number of recognizable classes, number of samples, metric expressions used. This paper discusses the possibility of transforming these networks in order to apply classical learning algorithms to them without using analytical expressions that calculate weight values. In the received network, training is carried out by recognizing images in pairs. This approach simplifies the learning process and easily allows to expand the neural network by adding new images to the recognition task. The advantages of these networks, including such as: 1) network architecture simplicity and transparency; 2) training simplicity and reliability; 3) the possibility of using a large number of images in the recognition problem using a neural network; 4) a consistent increase in the number of recognizable classes without changing the previous values of weights and thresholds.

**1.Introduction.**

At present, the challenge remains to create simple and clear architectures of neural networks with transparent architecture, capable of solving complex tasks. The current architectures of neural networks allow you to perform various tasks but have a number of limitations and weaknesses, such as:

1. The complexity of choosing and creating neural network architecture and its parameters. Neural network structures parameters are either too uncertain [1,2] or complex [3,4].

2. The complexity of training a neural network with large number of recognizable classes. The greater the number of recognizable classes, the harder neural networks are trained, and the worse the result of such training. Such limitation restricts the possible number of classes (N) used in recognition problems with neural networks application.

3. Another complication is the expansion of already trained neural network and the addition of new classes. The addition of new classes to the already trained network requires the retraining process with the full [1,2] or partial change of the previous parameters of neural network $\overline{W}$, $\overline{B}$ [3,4].

In this paper, we consider the intuitive, simple and transparent architecture of neural network, which allows circumventing the above-mentioned complications.

In the works [5,6], the architectures of neural networks implemented on the basis of metric recognition methods [7] were considered. In particular, Fig. 1 shows the scheme of feed forward architecture of three-layer neural network with threshold activation function that implements the nearest-neighbor method [3]. A set of selected examples (samples) of the training set was used as neighboring elements set. A set of samples can be selected both intuitively and by using the selection algorithm [6].

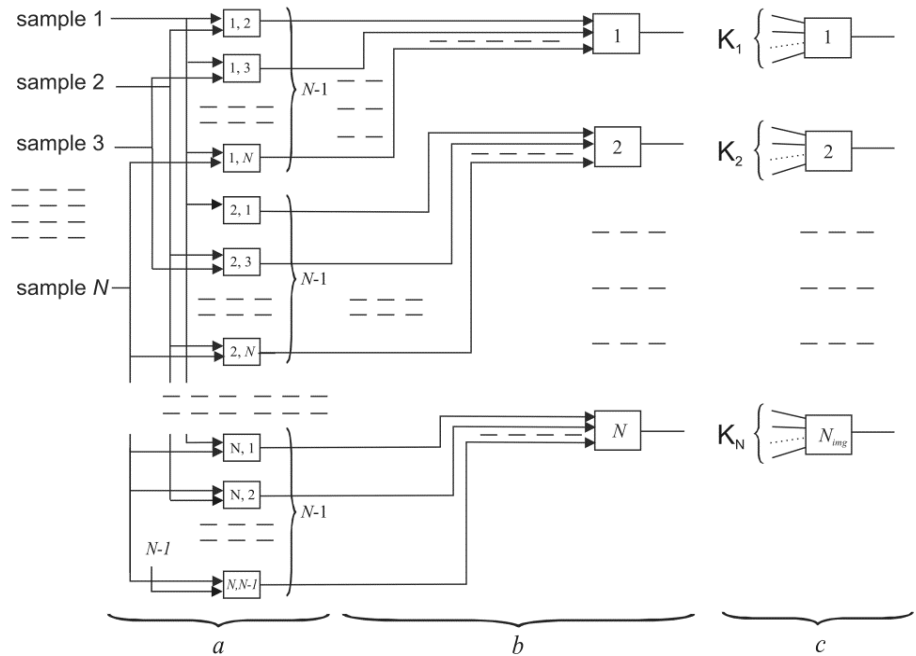

**Fig. 1** Network architecture based on the nearest-neighbor method for $N$ samples.

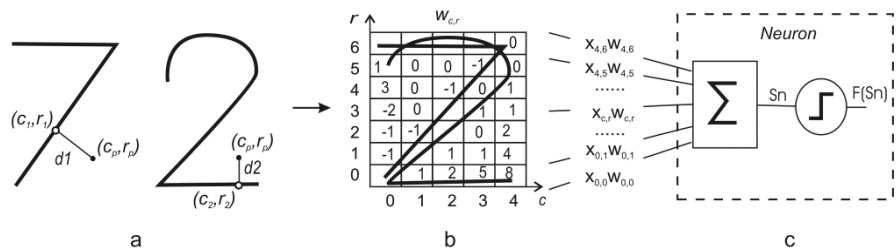

**Fig. 2** (a) Distances $d_1$ and $d_2$ for a point $(c_p, r_p)$, (b) Weight table for samples "*2*" and "*7*", (c) Neuron with threshold activation function.

The structure of this network (number of neurons, layers, and connections) is strictly determined according to the scheme shown in Fig.1. Each first-layer neuron performs a comparison of the images of two samples Fig.2. At that, the weights of first-layer neurons are determined analytically, based on metric expressions, for example, for Fig.2b by the expression:

$$w_{c,r}^{(1)} = d_1^2 - d_2^2 = \left((c_1 - c_p)^2 + (r_1 - r_p)^2\right) - \left((c_2 - c_p)^2 + (r_2 - r_p)^2\right), \quad (1)$$

wherein $(c_1, r_1)$ and $(c_2, r_2)$ are the closest points (cells) of sample images to the point (cell) $(c_p, r_p)$ (fig.2a).

The weighted sum $Sn_{i,j}^{(1)}$ and the activation function $f\left(Sn_{i,j}^{(1)}\right)$ for each first-layer neuron are determined by the expressions (2, 3):

$$Sn_{i,j}^{(1)} = \sum_{r=0}^{R}\sum_{c=0}^{C} x_{c,r} w_{c,r}^{(1)}, \qquad (2)$$

$$\begin{aligned} f\left(Sn_{i,j}^{(1)}\right) &= 1, \text{ if } Sn_{i,j}^{(1)} < 0 \\ f\left(Sn_{i,j}^{(1)}\right) &= 0, \text{ if } Sn_{i,j}^{(1)} > 0 \end{aligned}, \qquad (3)$$

wherein *C,R* are the quantity of columns and rows of the table (in fig.3b, *C=4, R=6*). The weights value for each input of second-layer and third-layer neuron is equal to 1.

$$w_{i,j}^{(2)} = w_{i,j}^{(3)} = 1, \qquad (4)$$

The number of the second-layer neurons is equal to the number of used samples $n_2=N$. The values of the weighted sum and the activation function for each *k*-th second-layer neuron are determined by the expressions (5, 6):

$$Sn_k^{(2)} = \sum_{j=1, j\neq k}^{N} f\left(Sn_{k,j}^{(1)}\right), \qquad (5)$$

$$\begin{aligned} f\left(Sn_k^{(2)}\right) &= 1, \text{ if } Sn_k^{(2)} \geq (N-1) = B^{(2)} \\ f\left(Sn_k^{(2)}\right) &= 0, \text{ if } Sn_k^{(2)} < (N-1) = B^{(2)} \end{aligned}, \qquad (6)$$

Here $B^{(2)} = N-1$ is the threshold value of the second-layer neuron. If the output of the *k*-th second-layer neuron equals $y_k^{(2)} = 1$, this would mean that the object at the input of the neural network have the corresponding *k*-th sample. Third-layer neurons combine the outputs of the samples of one *k*-th class into single output $y_k^{(3)}$. The weighted sum $Sn_k^{(3)}$ and the activation function $f\left(Sn_k^{(3)}\right)$ of the third-layer are determined by the expressions:

$$Sn_k^{(3)} = \sum_{i \in k}^{K} f\left(Sn_i^{(2)}\right), \qquad (7)$$

$$\begin{aligned} f\left(Sn_k^{(3)}\right) &= 1, \text{ if } Sn_k^{(3)} > 0 \\ f\left(Sn_k^{(3)}\right) &= 0, \text{ if } Sn_k^{(3)} = 0 \end{aligned}, \qquad (8)$$

If each recognizable class in Fig.1 has one corresponding sample, then, in this case, the neural network architecture is converted into a two-layer neural network, Fig.1ab, where the number of the second-layer neurons is equal to the number of recognizable classes *N*.

As already mentioned, the neural network architecture in Fig.1 implements the nearest-neighbor method, where the first layer weights value is calculated analytically (1) on the basis of the selected set of samples without using the classical training algorithms. In doing so, the question arises: 'Can this neural network architecture be trained by the classical training algorithms without using the selected samples and analytical expressions?'

## 2. Training of the neural network

In order to make the scheme in Fig.1ab trainable, let's take each first-layer neuron as a separate block of the neural network (*NN*<sub>i,j</sub>) (Fig.4). Fig.3a, performing the task of dividing the two images. At that, each *NN*<sub>i,j</sub> block has one binary output $y_{i,j}^{(1)} \in \{0, 1\}$, the value of which indicates the proximity of the object $\bar{X}$ to the image *i* or *j*. The first block *NN*<sub>1,2</sub> in Fig. 3a executes the division of the images *1* and *2*, and the second block *NN*<sub>1,3</sub> divides the images *1* and *3* and so on all the way the block *NN*<sub>1,N-1</sub>, separating the images *1* and *N-1* followed by *NN*<sub>i,j</sub>

blocks to divide the pairs of images {2, 1}, {2, 3}, ... {2, N-1}, {3, 1} and so on. Each $NN_{i,j}$ block is trained separately on the division of its pair of images. For $NN_{i,j}$ block training data only objects of *i, j* images are used Fig.3. Any known classical learning algorithms can be used for $NN_{i,j}$ block training. If object $\bar{X}$ belonging to the *k*-th image is used at the input of neural network in Fig.3a, then in the *N-1* outputs of the first layer would be active, starting from the block $NN_{k,j}$ and ending with the block $NN_{k,N}$, ($k \neq j$) Fig.3a. And according to (5, 6), the *k*-th neuron output in the second layer (Fig.3b) would be active ($y_k^{(2)} = f\left(Sn_k^{(2)}\right) = 1$), because $Sn_k^{(2)} = N - 1$. In the output of other second-layer neurons ($i \neq k$) the value for all $y_i^{(2)} = 0$, because for each $i \neq k$ there is one $NN_{i,k}$ unit in which $y_{i,k}^{(1)} = 0$. Thus, the maximum value $Max\left(Sn_{i \neq k}^{(2)}\right) = N - 2$, then from (6) it should be $f\left(Sn_{i \neq k}^{(2)}\right) = 0$.

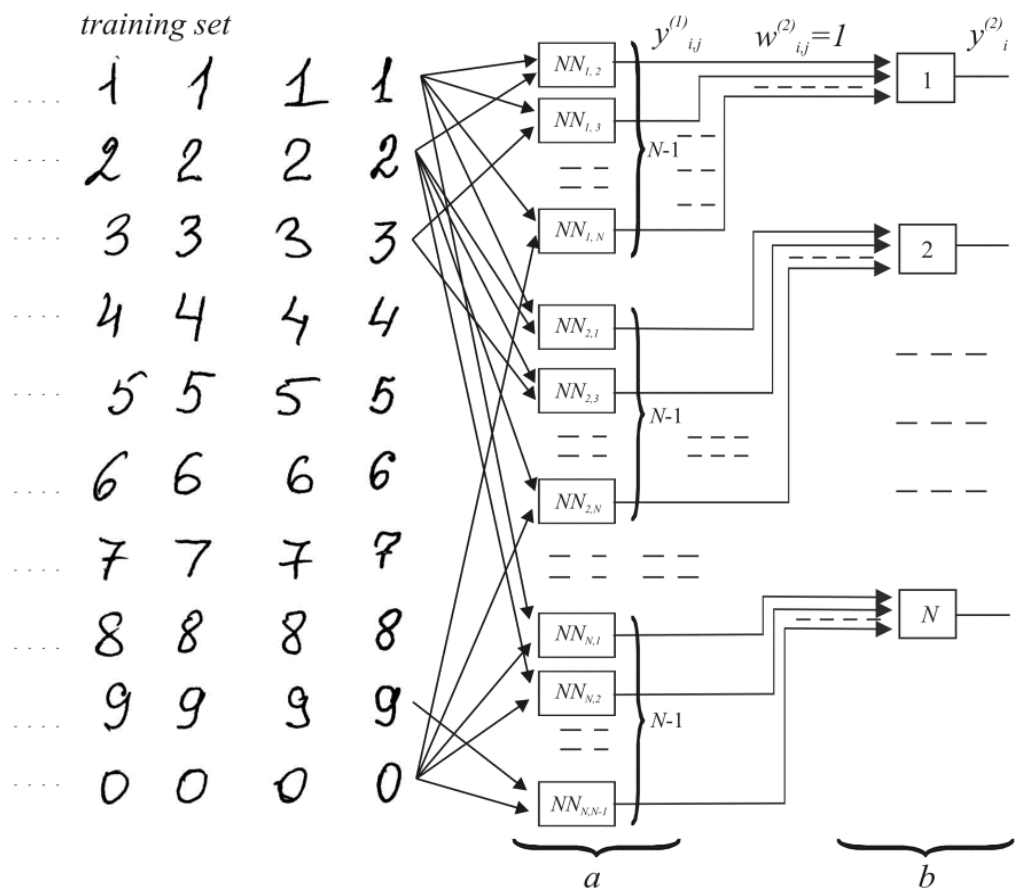

**Fig.3** Neural network training scheme.

In the simplest case, the $NN_{i,j}$ block can consist of one neuron. This will be the case where the divided images are distanced enough from each other and are easily divided by one hyperplane. In the more complex cases, each unit may consist of several neurons and layers. It is also possible to establish the fixed number of neurons and layers for each $NN_{i,j}$ block. $NN_{i,j}$ block neurons may have both threshold, and sigmoid activation function. The latter case would require an additional neuron at the output of the block $NN_{i,j}$, converting the output value of the block $NN_{i,j}$ into the binary value $y_{i,j}^{(1)} \in \{0, 1\}$, Fig.4. For this neuron, activation function will be determined by the expression (9). Any known classical algorithms may be used as the training algorithms for blocks $NN_{i,j}$.

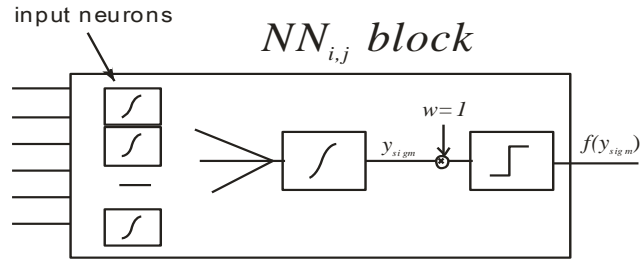

**Fig. 4** Conversion of output $NN_{i,j}$ into binary form.

$$f(y_{sigm}) = 1, if\ y_{sigm} > 0{,}5 \\ f(y_{sigm}) = 0, if\ y_{sigm} \leq 0{,}5 \quad , \tag{9}$$

If there are outputs of the second layer Fig.3b, which correspond to the same class of images Fig.5b, then the images of one class may be combined in a neuron of the third layer Fig.5c. For this neuron Fig.5c, the weighted sum and the activation function is determined by the expression (7, 8), and the values of all weights $w^{(3)}$ is also equal to 1.

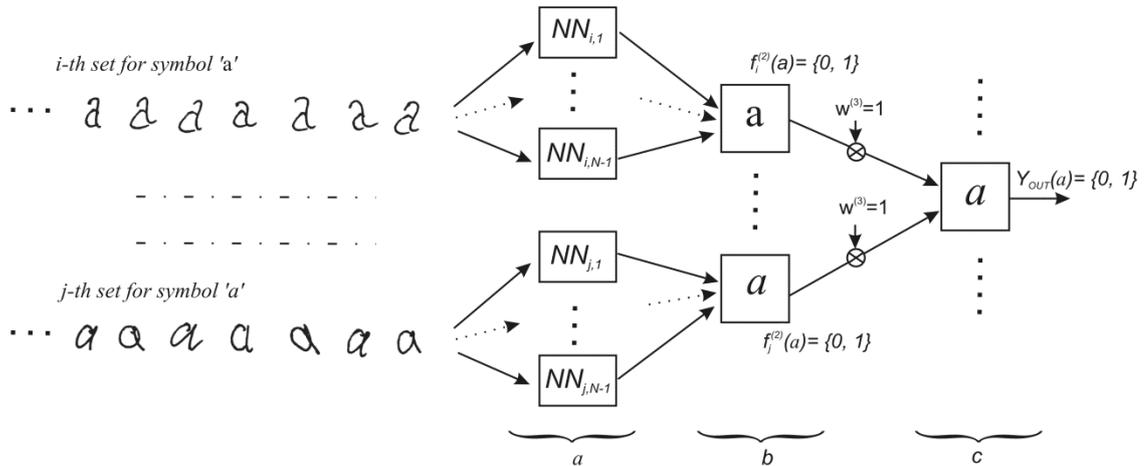

**Fig. 5.** Combining images of one class in a neuron of the third layer.

The disadvantage of the neural network architecture in Fig.3 is a large number of blocks $NN_{i,j}$, ($n_1$), which for the scheme in Fig.3 is increased by the formula:

$$n_1=(N-1)N, \tag{10}$$

Here it is possible to reduce this amount by half, if the blocks repeating division of similar pairs of images, such as {1, 2} and {2, 1} or {2, 3} and {3, 2} and so on [5] are excluded. In this case, the output of $NN_{i,j}$ block - $y_{i,j}^{(1)}$, performing the division of $i$ and $j$ images, will be connected to the $i$-th neuron of the second layer ($y_{i,j}^{(1)} = x_i^{(2)}$), and the inverted output value $y_{i,j}^{(1)}$ with the $j$-th neuron of the second layer $\bar{y}_{i,j}^{(1)} = x_j^{(2)}$ , Fig.6. We may assume that the number of $NN_{i,j}$ blocks may be even smaller, on the assumption that not all blocks $NN_{i,j}$ are vital and, perhaps, some of them may be deleted. The last statement requires additional studies.

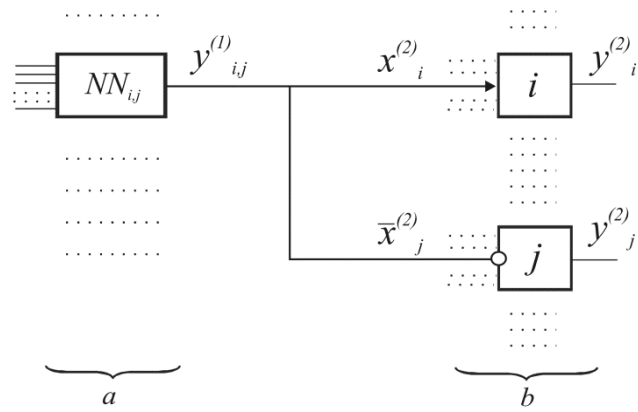

$$\underbrace{\phantom{XXXXXX}}_{a} \quad \underbrace{\phantom{XXXXXX}}_{b}$$

**Fig.6** Network compression by excluding $NN_{j,i}$ block.

Figure 7 shows the scheme of a compressed neural network for the problem of pattern recognition of three symbols: "A, B, C". The number of *NN* blocks is determined by expression:

$$n^{(1)} = C_N^2 = \frac{N!}{2!(N-2)!} = \frac{N(N-1)}{2} = \frac{3(3-1)}{2} = 3, \quad (11)$$

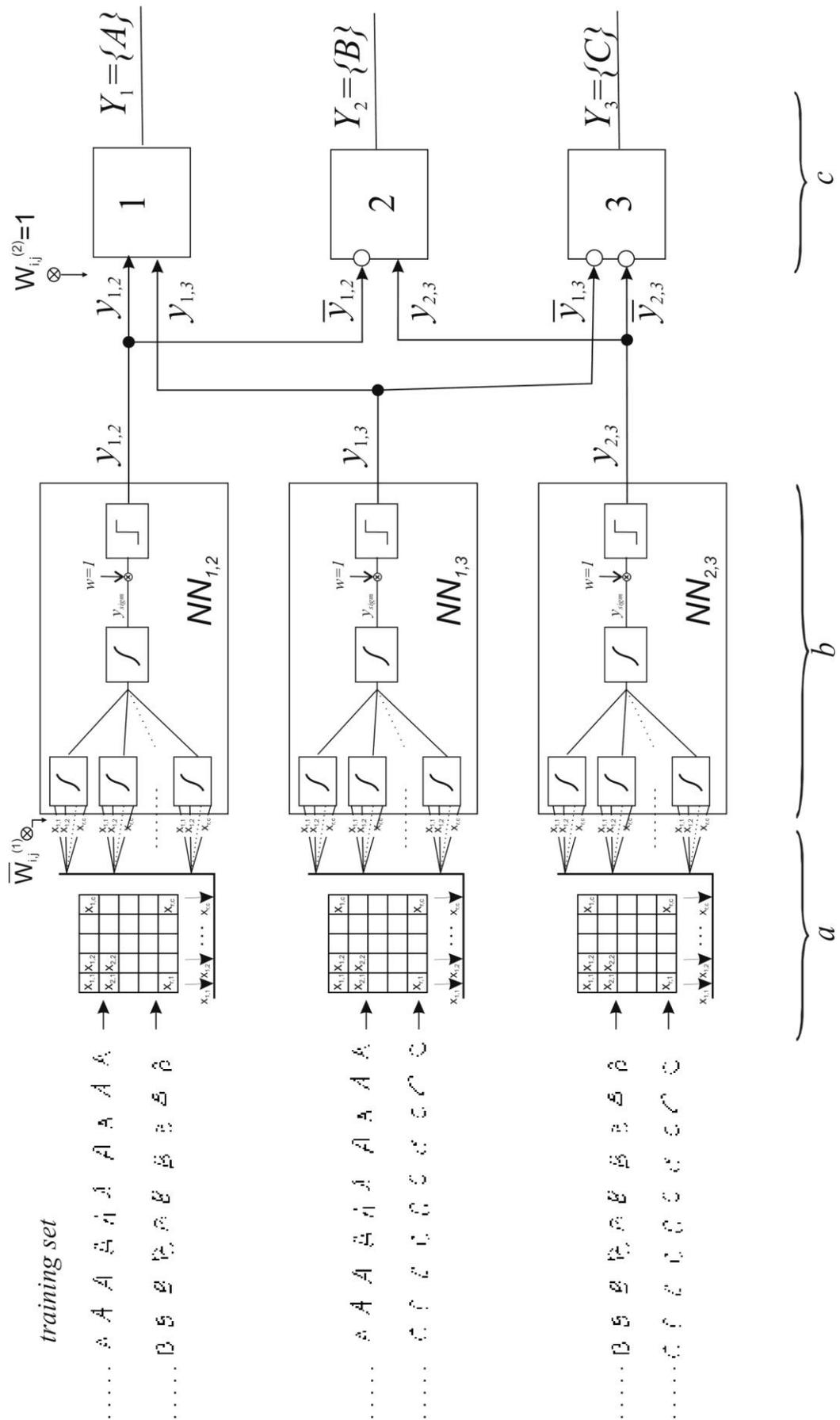

Fig. 7. Scheme of the neural network for the task of recognizing three images, (a) binary tables, (b) layer of blocks $NN_{ij}$ (c) the second layer of the neural network.

Thus, the neural network training in Fig.3 comes down to the training of each $NN_{i,j}$ block. Neural network training becomes easier, clearer and more reliable, since the training of $N$ images variety results in the problem of two images' paired division. Here there is also no need to retrain the whole network or the layers shown in Fig.3b, Fig.5c. Training of each $NN_{i,j}$ block is performed independent of the other $NN_{i,j}$ blocks, which allows us to easily expand the neural network by adding new recognizable images to the general problem. The possible number of recognizable images may be infinitely large ($N \to \infty$). For each new $N+1$ - th image of neural network in the scheme in Fig.3, one $NN_{k,N+1}$ block is added to each $k$-th image ($N$ blocks in total) and $N$ blocks $NN_{N+1,j}$ for $N+1$-th image. Total - $2N$ blocks $NN_{i,j}$, where $N$ is the last value of the images quantity (12). In the case of the compressed version of the neural network (Fig.6), for each new image only $N$ new blocks $NN_{i,j\ are}$ added (13). When adding the image, the threshold value $B^{(2)}$ of the second-layer neuron (6) Fig.3b increases by $1$.

$$n_{new\ neurons} = (N+1)((N+1)-1) - N(N-1) = N^2 + N - N^2 + N = 2N, \quad (12)$$

$$n_{new\ neurons} = (N+1)!/(2!((N+1)-1)!) - N!/2!(N-1)! = (N+1)((N+1)-1)/2 - N(N-1)/2 = (N^2+N-N^2+N)/2 = N. \quad (13)$$

### 3. Conclusion

Note that the considered architecture of the neural network has a number of advantages, and in particular:

1. In comparison with convolutional networks [3], the considered network architecture is simple.

2. Except for the dimension parameter of the binary matrix (fig.7a) and the number of input neurons in the NN block (fig.4, fig.7b), all other parameters of the network architecture are strictly determined from the initial conditions of the problem - the number of recognizable images (classes).

3. The number of recognizable images can easily and consistently increase and reach very large values, allowing to solve problems with a large number of recognizable classes. In addition, when adding new images (classes), the previous weight and threshold values do not change.

4. Learning uses classical training algorithms. Learning is simpler, as training separates the two images. It is clear that this requires a smaller amount of training examples and the number of epochs. For the same reason, it is almost impossible to hit the local minimum.

We can also add that the considered possibilities of the neural network architecture shown in Fig.3 have some similarity with the capabilities of biological brain. In particular, in the considered neural network architecture objects are recognized immediately as a whole, just as the biological brain would do. This distinguishes this architecture, for example, from the architecture of deep neural networks [3, 4], where objects are known to be sequentially analyzed and recognized from small image details to complex. In addition, in the architecture proposed in Fig. 3, 7, the number of recognizable images is almost unlimited, which is also very similar to the capabilities of the biological brain, which can easily recognize a huge number of images.
.